\documentclass[11pt,a4paper]{article}

\usepackage[utf8]{inputenc}
\usepackage[T1]{fontenc}
\usepackage{lmodern}
\usepackage{charter}
\usepackage[scaled=0.88]{beramono}
\usepackage{microtype}
\usepackage[margin=1in]{geometry}
\usepackage{hyperref}
\usepackage{url}
\usepackage{booktabs}
\usepackage{array}
\usepackage{colortbl}
\usepackage{multirow}
\usepackage{amsfonts}
\usepackage{amsmath}
\usepackage{xcolor}
\usepackage{graphicx}
\usepackage{listings}
\usepackage{titlesec}
\usepackage{parskip}
\usepackage{enumitem}
\usepackage{fancyhdr}
\usepackage{tikz}
\usepackage{pgf}
\usetikzlibrary{shapes.geometric, shapes.misc, arrows.meta, positioning, fit, backgrounds, calc}
\usepackage{subcaption}

\definecolor{cgNavy}{RGB}{24,48,84}
\definecolor{cgBlue}{RGB}{54,96,164}
\definecolor{cgTeal}{RGB}{38,122,145}
\definecolor{cgSoft}{RGB}{236,243,250}
\definecolor{cgLine}{RGB}{182,201,224}
\definecolor{cgRowA}{RGB}{244,248,253}
\definecolor{cgRowB}{RGB}{255,255,255}
\definecolor{cgGen} {RGB}{228,240,255}

\hypersetup{
  colorlinks=true,
  linkcolor=cgBlue,
  urlcolor=cgTeal,
  citecolor=cgBlue
}

\titleformat{\section}{\Large\bfseries}{\thesection}{1em}{}
\titleformat{\subsection}{\large\bfseries}{\thesubsection}{1em}{}

\begin{document}

\begin{center}
{\color{cgBlue}\rule{0.78\textwidth}{1.1pt}}\\[1.4em]
{\fontsize{24}{29}\selectfont\bfseries {\color{cgTeal}CreativeGame}\textcolor{cgNavy}{: Toward Mechanic-Aware Creative Game Generation.}}\\[1.5em]

\setlength{\fboxsep}{7pt}
\colorbox{cgSoft}{\textcolor{cgTeal}{\textbf{\large CreativeGame Team}}}\\[1.2em]

\renewcommand{\arraystretch}{1.22}
\begin{minipage}{0.78\textwidth}
\raggedright
\textcolor{cgBlue}{\textbf{Team Members (listed alphabetically):}} Hongnan Ma $\cdot$ Han Wang $\cdot$ Shenglin Wang $\cdot$ Tieyue Yin $\cdot$ Yiwei Shi $\cdot$ Yucong Huang\\[0.55em]
\textcolor{cgBlue}{\textbf{Team Leaders:}} Yingtian Zou $\cdot$ Muning Wen $\cdot$ Mengyue Yang\\[0.55em]
\textcolor{cgBlue}{\textbf{Institutions:}} University of Bristol $\cdot$ Shanghai Jiao Tong University $\cdot$ Shandong University $\cdot$ Nanjing University $\cdot$ Sreal AI\\[0.75em]
\textcolor{cgBlue}{\textbf{Project Page:}} \href{https://yiweishi-cn.github.io/CreativeEvolutionGame/index.html}{\textcolor{cgTeal}{yiweishi-cn.github.io/CreativeEvolutionGame}}
\end{minipage}\\[0.9em]
\textcolor{cgBlue!60}{\small April 2026}\\[1.0em]

{\color{cgLine}\rule{0.78\textwidth}{0.9pt}}\\[0.2em]
\end{center}

\section*{Abstract}

Large language models can generate plausible game code, but turning this capability into \emph{iterative creative improvement} remains difficult. In practice, single-shot generation often produces brittle runtime behavior, weak accumulation of experience across versions, and creativity scores that are too subjective to serve as reliable optimization signals. A further limitation is that mechanics are frequently treated only as post-hoc descriptions, rather than as explicit objects that can be planned, tracked, preserved, and evaluated during generation.

This report presents \textbf{CreativeGame}, a multi-agent system for iterative HTML5 game generation that addresses these issues through four coupled ideas: a proxy reward centered on programmatic signals rather than pure LLM judgment; lineage-scoped memory for cross-version experience accumulation; runtime validation integrated into both repair and reward; and a mechanic-guided planning loop in which retrieved mechanic knowledge is converted into an explicit mechanic plan before code generation begins. The goal is not merely to produce a playable artifact in one step, but to support interpretable version-to-version evolution.

The current system contains 71 stored lineages, 88 saved nodes, and a 774-entry global mechanic archive, implemented in 6{,}181 lines of Python together with inspection and visualization tooling. The system is therefore substantial enough to support architectural analysis, reward inspection, and real lineage-level case studies rather than only prompt-level demos.

A real 4-generation lineage shows that mechanic-level innovation can emerge in later versions and can be inspected directly through version-to-version records. The central contribution is therefore not only game generation, but a concrete pipeline for observing progressive evolution through explicit mechanic change.

\begin{center}
\rule{0.85\textwidth}{0.4pt}
\end{center}

\newpage

\tableofcontents

\newpage

\section{Introduction}
\label{sec:intro}

Generating creative interactive content (games) remains an unsolved problem for LLMs. A single LLM call given ``make me a creative game'' produces plausible-looking code that often fails at runtime, generic templates (Pong clones, basic shooters), no mechanism for accumulating what worked across generations, and subjective creativity scores that are hard to validate. More broadly, creativity research has long emphasized both the difficulty of judging creativity and the importance of evaluative perspective \cite{KaufmanBaer2012,GlaveanuBeghetto2021}. The fundamental difficulty in our setting is therefore that \textbf{creativity is open-ended, hard to evaluate, and hard to optimize}, yet a measurable optimization signal is required for iterative improvement.

CreativeGame addresses this problem through five tightly coupled design choices. First, creative game generation is decomposed into 7 logical agents (10 executable roles including 4 code-generation sub-agents), each with a focused prompt and parameter profile. Second, subjective LLM scoring is replaced with a \textbf{CreativeProxyReward} composed of weighted proxy signals and gating conditions, so that only a minority of the total signal depends directly on LLM judgment. Third, the system introduces a \textbf{lineage-aware memory} in which all forks within a lineage share a common learned memory pool, allowing experience to accumulate across versions; this design is directly informed by recent MemRL-style work on runtime reinforcement learning over episodic memory \cite{zhang2026memrlselfevolvingagentsruntime}. Fourth, a \textbf{runtime validator} performs deep static analysis and optional browser execution to catch bugs that model-side evaluation may miss. Fifth, \textbf{mechanics are promoted to explicit planning objects}: planner-time mechanic retrieval produces a structured mechanic plan that can later be compared with realized mechanics and stored in evolution records. Each generation call runs up to 3 total iterations (1 initial generation followed by up to 2 refinement passes), and only the final state is saved as a lineage node. This framing is important because the project is intended to support not only generation, but interpretable version-to-version evolution.

\subsection{Contributions}

\begin{enumerate}[leftmargin=1.5em]
\item A formulation of iterative game generation in which mechanics are treated as explicit planning and evaluation objects, rather than only as retrospective descriptions of generated content (Sections~\ref{sec:notation} and~\ref{sec:mechanic_guided}).
\item A \emph{CreativeProxyReward} whose dominant signals are deterministic Python-side measurements, combining mechanic realization, structural change, novelty, and runtime validation while reducing dependence on unconstrained LLM judgment (Section~\ref{sec:reward}).
\item A lineage-aware memory architecture that shares experience across versions within a lineage while preserving isolation across lineages, making iterative accumulation possible without collapsing all generations into a single global memory pool (Section~\ref{sec:memory}).
\item Integration of runtime validation directly into the generation loop as both a repair trigger and a reward gate, including a lightweight static analyzer and optional browser-based execution checks (Section~\ref{sec:tester}).
\item A fully self-contained implementation of the complete pipeline, together with real-lineage evidence showing concrete mechanic-level innovation across iterations (Sections~\ref{sec:implementation} and~\ref{sec:case}).
\end{enumerate}

\subsection{Report Scope}

This report describes the current CreativeGame system. Architectural descriptions and system statistics are aligned with the implemented pipeline and its stored generation data.

\section{Formal Foundations and Notation}
\label{sec:notation}

This report adopts a notation system aligned with the current project concept documents on game definition and mechanic definition. The aim is to keep the report, the code interpretation, and future evaluation criteria consistent.

\subsection{Game as a Rule-Organized Interactive System}

We represent a game as
\begin{equation}
G = (P, S, A, T, O, F, K, W, U, \Phi, C, R, M),
\label{eq:game_tuple}
\end{equation}
where $P$ denotes decision-bearing agents, $S$ the state space, $A$ the action space, $T$ the transition rule, $O$ the observation structure, $F$ the feedback mapping, $K$ the resource/constraint structure, $W$ the outcome structure, $U$ the preference ordering over outcomes, $\Phi$ the challenge structure, $C$ the content layer, $R$ the representation layer, and $M$ the meta layer.

Following the project definition, we distinguish:
\begin{equation}
G_{\text{core}} = (P, S, A, T, O, F, K, W, U, \Phi),
\label{eq:game_core}
\end{equation}
\begin{equation}
G_{\text{support}} = (C, R, M).
\label{eq:game_support}
\end{equation}
Changes to $G_{\text{core}}$ count as structural game changes, while changes confined to $G_{\text{support}}$ are cosmetic or presentation-level changes. This distinction is important throughout the report: the CreativeGame pipeline is intended to reward structural variation rather than pure reskinning.

\subsection{Meaningful Play and Learnability}

The concept documents define two validity predicates over the assembled game system. This emphasis on meaningful play is broadly compatible with classic game-design accounts that center rules, player interpretation, and consequence \cite{SalenZimmerman2003,Schell2008}.
\begin{equation}
\Psi(G) \in \{0,1\},
\label{eq:psi}
\end{equation}
\begin{equation}
\Lambda(G) \in \{0,1\},
\label{eq:lambda}
\end{equation}
where $\Psi(G)$ is the meaningful-play condition and $\Lambda(G)$ is the learnability condition. Intuitively, $\Psi(G)=1$ means that action outcomes are both discernible and integrated into the broader system, while $\Lambda(G)=1$ means that the game contains exploitable regularities such that non-random strategy can improve expected outcomes.

Accordingly, a valid game is one whose core structure is complete and which satisfies both predicates:
\begin{equation}
G \text{ is a valid game}
\iff
G_{\text{core}} \text{ is structurally complete}
\wedge \Psi(G)=1
\wedge \Lambda(G)=1.
\label{eq:valid_game}
\end{equation}

\subsection{Mechanic as a Local Rule Structure}

Within this report, a game mechanic is not treated as a theme tag or content label, but as a local rule structure inside the game system. This is broadly compatible with direct attempts to define game mechanics in terms of player interaction methods and rule-bearing game structures \cite{sicart2008defining}. We use the compact formalization
\begin{equation}
m = (\Delta A, \Delta T, \Delta O, \Delta F, \Delta K, \Delta W),
\label{eq:mechanic_tuple}
\end{equation}
meaning that a mechanic is identified by the stable way it changes at least one of the structural layers most relevant to play: action space, transition logic, information structure, feedback relation, resource constraints, or goal progression.

To distinguish existence from quality, we also use three mechanic-level scores:
\begin{equation}
m \mapsto (E_m, I_m, V_m),
\label{eq:mechanic_scores}
\end{equation}
where $E_m$ denotes mechanic existence, $I_m$ mechanic importance, and $V_m$ showcase value. In the concept documents, $E_m$ acts as a gate for whether a candidate should count as a mechanic at all, $I_m$ captures how strongly it matters to decision structure and the core loop, and $V_m$ captures whether the mechanic can be clearly observed in a short evaluation window.

\subsection{Structural Creativity and Mechanic Delta}

Given a parent game $G$ and a generated variant $G'$, structural creativity should ideally be attributed to changes in the core rule-bearing structure rather than to support-layer variation alone. We therefore use the notion of mechanic delta in the following sense:
\begin{equation}
\delta(G, G') = \{m \mid m \in G'_{\text{core}} \setminus G_{\text{core}}\}
\cup
\{m \mid m \in G_{\text{core}} \setminus G'_{\text{core}}\}.
\label{eq:mechanic_delta}
\end{equation}
In implementation, the live system approximates this idealized delta through extracted mechanic sets and related reward terms. Nevertheless, Equations~\ref{eq:game_tuple}--\ref{eq:mechanic_delta} provide the formal reference used throughout the report whenever it discusses structural change, mechanic preservation, novelty, and planned-vs-realized mechanics.

\section{System Architecture}
\label{sec:system}

\begin{figure}[t]
\centering
\begin{tikzpicture}[
  pbox/.style={rectangle, rounded corners=4pt,
               draw=cgBlue!70, fill=cgSoft,
               text width=8.6cm, align=center,
               font=\small, inner sep=7pt, minimum height=0.75cm},
  sbox/.style={pbox, fill=white, draw=cgLine, text width=7.8cm, font=\footnotesize},
  arr/.style={-{Stealth[length=5pt]}, thick, color=cgBlue!70},
  lbl/.style={font=\footnotesize\itshape, color=cgTeal}
]
\node[pbox, fill=cgNavy!8, draw=cgNavy!50]
  (prompt) {\textbf{\color{cgNavy}Prompt + Parent Code Context}};

\node[pbox, below=0.28cm of prompt]
  (planner) {\textbf{\color{cgBlue}Planner}\\[2pt]
  retrieves lineage memory $\cdot$ global memory $\cdot$ mechanic archive\\
  emits plan + \texttt{CURRENT\_MECHANIC\_SET}};

\node[pbox, below=0.28cm of planner]
  (gen) {\textbf{\color{cgBlue}Code Generation}\\[2pt]
  iteration 1:~Skeleton $\to$ Feature $\to$ Visual Enhancer\\
  later iterations:~Refinement};

\node[pbox, below=0.28cm of gen]
  (val) {\textbf{\color{cgBlue}Validation \& Repair}\\[2pt]
  structural check $\to$ runtime test $\to$ fixer (if needed)};

\node[pbox, below=0.28cm of val]
  (eval) {\textbf{\color{cgBlue}Evaluation}\\[2pt]
  scores $\cdot$ mechanic realization $\cdot$ mechanic delta $\cdot$ novelty grounding};

\node[pbox, below=0.28cm of eval]
  (refl) {\textbf{\color{cgBlue}Reflection}\\[2pt]
  verdict $\cdot$ memory payload $\cdot$ reward signal};

\node[pbox, below=0.28cm of refl, draw=cgTeal!60, fill=cgTeal!6]
  (loop) {\textbf{\color{cgTeal}Loop Control}\\[2pt]
  \textsc{continue} $\Rightarrow$ next iteration\quad
  \textsc{stop} $\Rightarrow$ output formatting + lineage save};

\foreach \a/\b in {prompt/planner, planner/gen, gen/val, val/eval, eval/refl, refl/loop}
  \draw[arr] (\a.south) -- (\b.north);

\draw[arr, dashed, color=cgTeal!70, rounded corners=6pt]
  (loop.east) -- ++(0.6,0) -- ++(0,3.6) -- node[lbl, right]{up to 3 iters} ++(0,3.6) -- (gen.east);
\end{tikzpicture}
\caption{Code-grounded overview of the implemented pipeline (\texttt{pipeline.py}, \texttt{agents.py}). The dashed feedback arc indicates the refinement loop: after \textsc{continue}, control returns to the Code Generation stage for up to 2 further passes before \textsc{stop} triggers output formatting and lineage save.}
\label{fig:pipeline}
\end{figure}
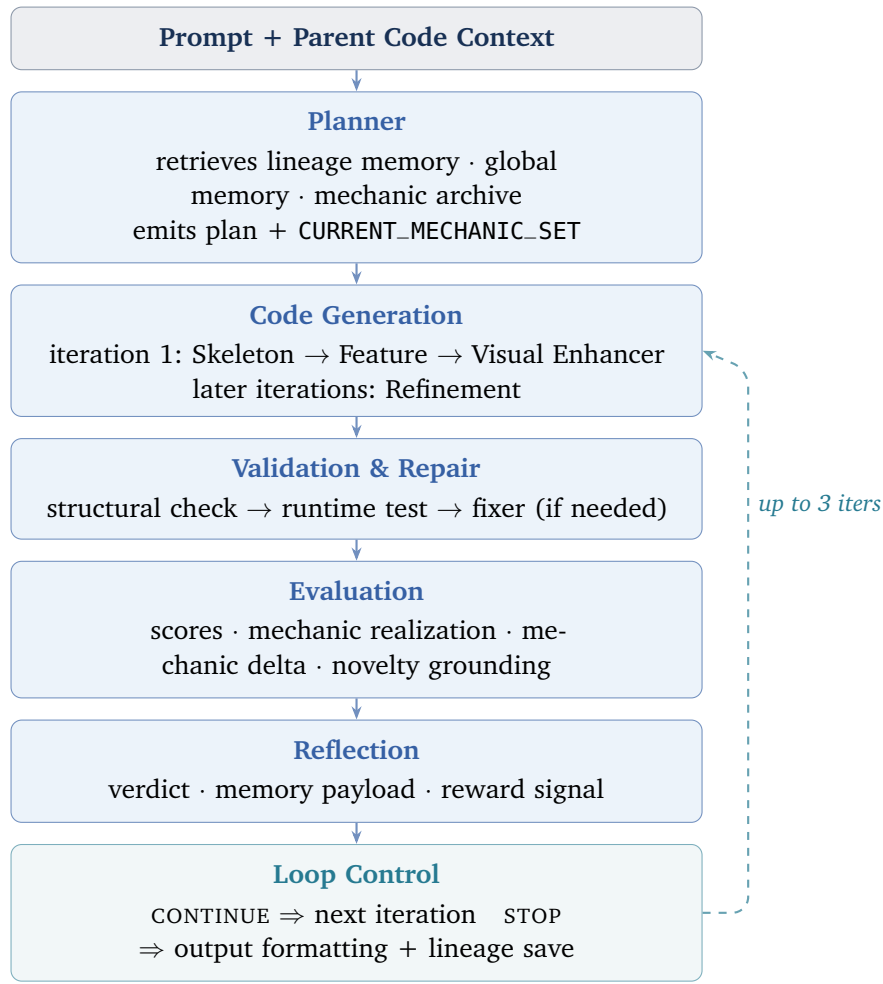

The system contains 7 logical agents, with the generation stage composed of 4 internal sub-stages (Skeleton, Feature, Visual, Refinement), yielding 10 distinct executable roles. Table~\ref{tab:agents} summarizes the configuration.

\begin{table}[htbp]
\caption{Role specifications in the current system. Generation parameters and
token budgets are tuned per role; the Generation stage is split into four
sequential sub-roles.}
\label{tab:agents}
\centering
\small
\setlength{\tabcolsep}{9pt}
\renewcommand{\arraystretch}{1.32}
\begin{tabular}{%
  >{\bfseries}l
  l
  >{\centering\arraybackslash}p{1.6cm}
  >{\raggedleft\arraybackslash}p{2.2cm}}
\rowcolor{cgNavy}
\color{white}\normalfont Role &
\color{white}Sub-role &
\color{white}Temp. &
\color{white}Token budget \\[0.5pt]
\rowcolor{cgRowA}
Planning           & —           & 0.7       & 12{,}000 \\
\rowcolor{cgGen}
\multirow{4}{*}{Generation}
                   & Skeleton    & 0.7       & \phantom{0}4{,}096 \\
\rowcolor{cgGen}
                   & Feature     & 0.8       & 16{,}000 \\
\rowcolor{cgGen}
                   & Visual      & 0.8       & 20{,}000 \\
\rowcolor{cgGen}
                   & Refinement  & 0.7       & 24{,}000 \\
\rowcolor{cgRowA}
Repair             & —           & 0.3       & 20{,}000 \\
\rowcolor{cgRowB}
Runtime Validation & —           & \textit{n/a} & \textit{n/a} \\
\rowcolor{cgRowA}
Evaluation         & —           & 0.2       & \phantom{0}4{,}000 \\
\rowcolor{cgRowB}
Reflection         & —           & 0.3       & \phantom{0}3{,}000 \\
\rowcolor{cgRowA}
Output Formatting  & —           & 0.2       & \phantom{0}5{,}000 \\
\bottomrule
\end{tabular}
\end{table}

\subsection{Reliability Mechanisms}

Three layers of error recovery are used: repeated model calls, stage-wise fallback, and tolerant final formatting. Together these reduced the pipeline failure rate from $\sim$10\% to $<$2\%.

\subsection{Iterations vs.\ Lineage Versions}

Each generation call runs up to 3 total iterations: 1 initial generation followed by up to 2 refinement passes. Only the final state is saved as a lineage node. User-visible ``v1/v2/v3/v4'' labels refer to \emph{separate generation calls}, not internal refinement passes within one call.

\subsection{Mechanic-Guided Planning Loop}
\label{sec:mechanic_guided}

\begin{figure}[t]
\centering
\begin{tikzpicture}[
  mbox/.style={rectangle, rounded corners=4pt,
               draw=cgBlue!60, fill=cgSoft,
               text width=7.8cm, align=center,
               font=\small, inner sep=6pt},
  abox/.style={mbox, fill=cgTeal!8, draw=cgTeal!60},
  arr/.style={-{Stealth[length=5pt]}, thick, color=cgBlue!70},
  warr/.style={-{Stealth[length=5pt]}, thick, dashed, color=cgTeal!80},
  lbl/.style={font=\footnotesize\itshape, color=cgTeal, midway, right=2pt}
]
\node[abox]
  (archive) {\textbf{\color{cgTeal}Global Mechanic Archive — Planner Query}\\[2pt]
  relevant $\cdot$ underexplored $\cdot$ overused $\cdot$ forbidden patterns};

\node[mbox, below=0.3cm of archive]
  (plan) {\textbf{\color{cgBlue}Planner Output}\\[2pt]
  \texttt{CURRENT\_MECHANIC\_SET}: preserve / add / remove / recombine};

\node[mbox, below=0.3cm of plan]
  (codegen) {\textbf{\color{cgBlue}Generation Stages}\\[2pt]
  mechanic contract prepended to Skeleton $\cdot$ Feature $\cdot$ Visual $\cdot$ Refinement};

\node[mbox, below=0.3cm of codegen]
  (evalout) {\textbf{\color{cgBlue}Evaluator Output}\\[2pt]
  realized mechanics $\cdot$ mechanic judgments $\cdot$ delta $\cdot$ novelty grounding};

\node[mbox, below=0.3cm of evalout]
  (refl) {\textbf{\color{cgBlue}Reflection + Memory Update}\\[2pt]
  reward-tagged experience $\cdot$ successful patterns $\cdot$ failed patterns};

\foreach \a/\b in {archive/plan, plan/codegen, codegen/evalout, evalout/refl}
  \draw[arr] (\a.south) -- (\b.north);

\draw[warr, rounded corners=8pt]
  (refl.east) -- ++(1.1,0) -- node[lbl]{write-back} ++(0,5.4) -- (archive.east);
\end{tikzpicture}
\caption{Mechanic-centered feedback loop. Mechanics are retrieved \emph{before} planning (top), converted into an explicit generation contract, compared against realized mechanics in evaluation, and conditionally written back into the archive (dashed arc) after reflection.}
\label{fig:mechanic_loop}
\end{figure}
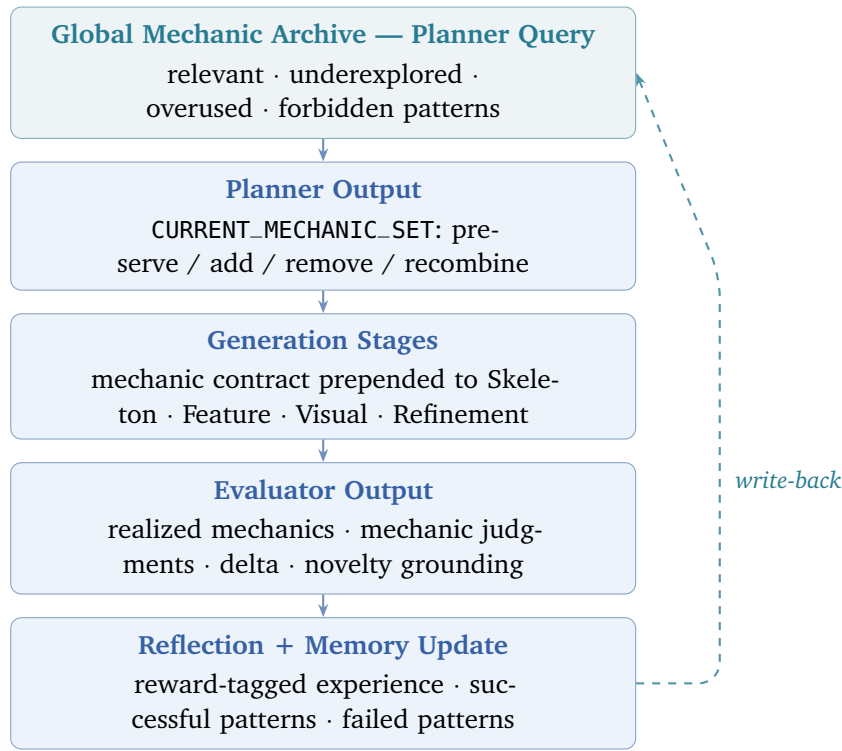

The system includes an explicit mechanic layer between prompt interpretation and code generation. Concretely, the planner receives retrieved mechanic-library context, emits a labeled mechanic plan, and the orchestration loop stores that structure for each iteration. The same plan is then appended to later evaluation and reflection stages, enabling planned-vs-realized mechanic comparison and mechanic-aware memory writing.

This changes the interpretation of the system in an important way. Mechanics are treated not merely as archive entries or post-hoc descriptors, but as explicit control variables: the planner can state which mechanics should be preserved, added, removed, or recombined before code generation begins. In the notation of Section~\ref{sec:notation}, the pipeline therefore moves beyond post-hoc content description toward explicit planning over candidate local rule structures $m$ and their intended changes to $G_{\text{core}}$. The system also exposes planning and evaluation records in an inspectable form, making version-to-version mechanic change directly observable.

\section{CreativeProxyReward}
\label{sec:reward}

Prior work uses LLM-based scoring as the primary reward signal. We observe three problems: (i)~score saturation (GPT-class models default to 7/10 regardless of input), (ii)~no verifiable improvement (a 7$\to$8 change is not statistically distinguishable from noise), and (iii)~Goodhart's-law risk (optimizing for LLM judgment leads to outputs that ``sound creative'' without being mechanically novel).

\subsection{Formula}

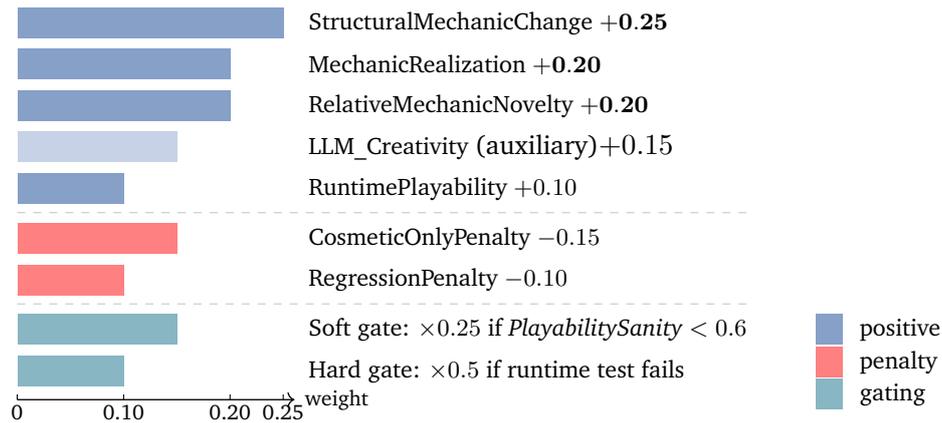
\begin{figure}[htbp]
\centering
\begin{tikzpicture}[
  bar/.style ={rectangle, anchor=west, minimum height=0.40cm,
               fill=cgBlue!60, draw=none},
  neg/.style ={bar, fill=red!50},
  gate/.style={bar, fill=cgTeal!55},
  lbl/.style ={font=\footnotesize, anchor=west},
]
\node[bar,  minimum width=3.50cm] (b1) at (0, 0.00) {};
\node[lbl] at (3.70, 0.00) {StructuralMechanicChange \hfill $\mathbf{+0.25}$};

\node[bar,  minimum width=2.80cm] (b2) at (0,-0.55) {};
\node[lbl] at (3.70,-0.55) {MechanicRealization \hfill $\mathbf{+0.20}$};

\node[bar,  minimum width=2.80cm] (b3) at (0,-1.10) {};
\node[lbl] at (3.70,-1.10) {RelativeMechanicNovelty \hfill $\mathbf{+0.20}$};

\node[bar,  minimum width=2.10cm, fill=cgBlue!28] (b4) at (0,-1.65) {};
\node[lbl] at (3.70,-1.65) {LLM\_Creativity \small(auxiliary)\normalsize \hfill $+0.15$};

\node[bar,  minimum width=1.40cm] (b5) at (0,-2.20) {};
\node[lbl] at (3.70,-2.20) {RuntimePlayability \hfill $+0.10$};

\draw[dashed, color=gray!45, line width=0.4pt] (0,-2.52) -- (9.6,-2.52);

\node[neg,  minimum width=2.10cm] (b6) at (0,-2.86) {};
\node[lbl] at (3.70,-2.86) {CosmeticOnlyPenalty \hfill $-0.15$};

\node[neg,  minimum width=1.40cm] (b7) at (0,-3.41) {};
\node[lbl] at (3.70,-3.41) {RegressionPenalty \hfill $-0.10$};

\draw[dashed, color=gray!45, line width=0.4pt] (0,-3.73) -- (9.6,-3.73);

\node[gate, minimum width=2.10cm] (g1) at (0,-4.07) {};
\node[lbl] at (3.70,-4.07)
  {Soft gate: $\times 0.25$ if \textit{PlayabilitySanity} $< 0.6$};

\node[gate, minimum width=1.40cm] (g2) at (0,-4.62) {};
\node[lbl] at (3.70,-4.62)
  {Hard gate: $\times 0.5$ if runtime test fails};

\draw[->, line width=0.5pt] (0,-5.00) -- (3.65,-5.00)
  node[right, font=\scriptsize]{weight};
\foreach \x/\t in {0/0, 1.40/0.10, 2.80/0.20, 3.50/0.25}
  \draw[line width=0.4pt] (\x,-5.00) -- (\x,-4.93)
    node[below, font=\scriptsize]{\t};

\node[bar,  minimum width=0.35cm] at (10.50,-4.07) {};
\node[font=\footnotesize, anchor=west] at (10.95,-4.07) {positive};
\node[neg,  minimum width=0.35cm] at (10.50,-4.50) {};
\node[font=\footnotesize, anchor=west] at (10.95,-4.50) {penalty};
\node[gate, minimum width=0.35cm] at (10.50,-4.93) {};
\node[font=\footnotesize, anchor=west] at (10.95,-4.93) {gating};
\end{tikzpicture}
\caption{CreativeProxyReward signal weights (scale: 3.5\,cm\,$= 0.25$). The three
mechanic-grounded signals account for 65\% of the maximum positive weight; LLM
judgment contributes only 15\%. Gating conditions are applied multiplicatively
after summing the weighted terms.}
\label{fig:reward_bars}
\end{figure}

The CreativeProxyReward consists of 7 weighted signal terms and 2 gating conditions:
\begin{equation}
\label{eq:reward}
\begin{aligned}
\textit{Reward} =\;& +0.20 \cdot \textit{MechanicRealization} \\
&+0.25 \cdot \textit{StructuralMechanicChange} \\
&+0.20 \cdot \textit{RelativeMechanicNovelty} \\
&+0.15 \cdot \textit{LLM\_Creativity} \\
&+0.10 \cdot \textit{RuntimePlayability} \\
&-0.15 \cdot \textit{CosmeticOnlyPenalty} \\
&-0.10 \cdot \textit{RegressionPenalty}
\end{aligned}
\end{equation}
subject to two gating conditions (Figure~\ref{fig:reward_bars}):
\begin{equation}
\textit{Reward} \leftarrow
\begin{cases}
0.25 \cdot \textit{Reward} & \text{if } \textit{PlayabilitySanity} < 0.6 \quad\text{(soft gate)}\\
0.5 \cdot \textit{Reward} & \text{if not } \textit{runtime\_test\_passed} \quad\text{(hard gate)}
\end{cases}
\end{equation}

\subsection{Signal Sources}

\textit{MechanicRealization} measures whether the generated game actually realizes the planned mechanics. In the formal view, this acts as an implementation proxy for whether the intended mechanic-level changes to $G_{\text{core}}$ are realized in the generated artifact. \textit{StructuralMechanicChange} is computed from added, modified, and removed mechanics together with an overall structural-change score. \textit{RelativeMechanicNovelty} is grounded against the global mechanic archive, which currently contains 774 entries. \textit{LLM\_Creativity} is $(\text{score}-3)/7$ clamped to $[0,1]$. \textit{RuntimePlayability} is the tester score (Section~\ref{sec:tester}); despite the historical field name, we interpret it here as a proxy for runtime robustness and execution quality rather than a direct measure of human-perceived fun. \textit{CosmeticOnlyPenalty} penalizes outputs with negligible structural change, and \textit{RegressionPenalty} captures missing executable core features such as canvas setup, game loop, or input handling.

\subsection{Relation to the Formal Notation}

The current reward implementation should be understood as an engineering proxy rather than a complete realization of the formal framework in Section~\ref{sec:notation}. In the formal view, structural creativity should respond to changes in $G_{\text{core}}$, to the appearance or preservation of local rule structures $m$, and ideally to their contribution to meaningful play $\Psi(G)$ and learnability $\Lambda(G)$. The present implementation approximates this target through extracted mechanic deltas, mechanic realization, archive novelty, runtime robustness, and penalties for purely cosmetic or regressive outputs. The distinction matters: the notation defines the target semantics, whereas the current implementation provides an operational approximation.

\subsection{Why the Hard Gate}

The runtime hard gate (\textit{Reward} $\times$ 0.5 if test fails) is critical because it prevents the system from rewarding ``creative-looking'' games that do not actually run. Unlike LLM-based signals, this signal cannot be gamed by LLM optimization because it executes the actual code. In the current formulation, \textit{LLM\_Creativity} remains auxiliary, while mechanic realization, structural change, novelty, and runtime robustness dominate the score.

\section{Lineage-Aware Memory}
\label{sec:memory}

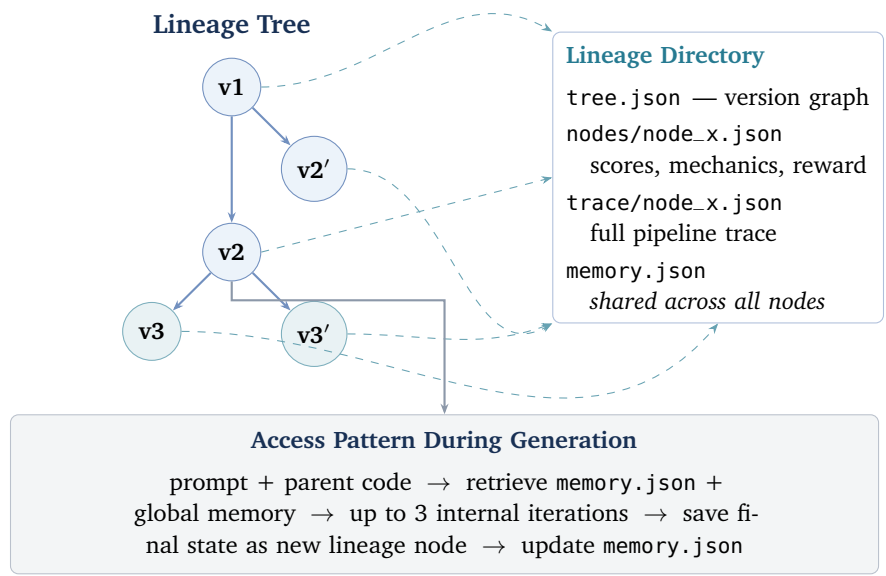
\begin{figure}[htbp]
\centering
\begin{tikzpicture}[
  vnode/.style={circle, draw=cgBlue!70, fill=cgSoft,
                minimum size=0.72cm, font=\footnotesize\bfseries},
  leaf/.style={vnode, fill=cgTeal!10, draw=cgTeal!60},
  mem/.style={rectangle, rounded corners=3pt, draw=cgBlue!40, fill=white,
              text width=4.0cm, align=left, font=\footnotesize, inner sep=5pt},
  acc/.style={rectangle, rounded corners=3pt, draw=cgNavy!30, fill=cgNavy!5,
              text width=11.0cm, align=center, font=\footnotesize, inner sep=6pt},
  dirarr/.style={-{Stealth[length=4pt]}, thick, color=cgBlue!70},
  memarr/.style={-{Stealth[length=4pt]}, dashed, color=cgTeal!70},
]

\node[vnode]                         (v1)  at (0,  0)    {v1};
\node[vnode, below=0.6cm of v1]      (v2a) at (0, -1.2)  {v2};
\node[leaf,  below left =0.5cm and 0.5cm of v2a] (v3a) {v3};
\node[leaf,  below right=0.5cm and 0.5cm of v2a] (v3b) {v3$'$};
\node[vnode, below right=0.5cm and 0.5cm of v1]  (v2b) {v2$'$};

\draw[dirarr] (v1)  -- (v2a);
\draw[dirarr] (v1)  -- (v2b);
\draw[dirarr] (v2a) -- (v3a);
\draw[dirarr] (v2a) -- (v3b);

\node[above=0.12cm of v1, font=\small\bfseries\color{cgNavy}]
  {Lineage Tree};

\node[mem] (membox) at (6.4, -1.2)
  {\textbf{\color{cgTeal}Lineage Directory}\\[4pt]
   \texttt{tree.json}\enspace— version graph\\[2pt]
   \texttt{nodes/node\_x.json}\\
   \quad scores, mechanics, reward\\[2pt]
   \texttt{trace/node\_x.json}\\
   \quad full pipeline trace\\[2pt]
   \texttt{memory.json}\\
   \quad \textit{shared across all nodes}};

\draw[memarr] (v1.east)  to[out=0,  in=150] (membox.north west);
\draw[memarr] (v2a.east) --                  (membox.west);
\draw[memarr] (v2b.east) to[out=0,  in=210] (membox.south west);
\draw[memarr] (v3a.east) to[out=0,  in=230] (membox.south);
\draw[memarr] (v3b.east) to[out=0,  in=200] (membox.south west);

\node[acc] (access) at (2.8, -5.4)
  {\textbf{\color{cgNavy}Access Pattern During Generation}\\[4pt]
   prompt + parent code\;
   $\to$\; retrieve \texttt{memory.json} + global memory\;
   $\to$\; up to 3 internal iterations\;
   $\to$\; save final state as new lineage node\;
   $\to$\; update \texttt{memory.json}};

\draw[dirarr, color=cgNavy!50]
  (v2a.south) -- ++(0,-0.25) -| (access.north);

\end{tikzpicture}
\caption{Lineage-level storage and memory sharing (\texttt{memory/manager.py}). All
nodes share \texttt{memory.json} (dashed arcs); lineages are isolated from each
other. Internal refinement iterations occur \emph{inside} one generation call;
user-visible versions (v1, v2, \ldots) are stored as separate tree nodes.}
\label{fig:lineage_memory}
\end{figure}

We organize game versions as a \textbf{lineage tree}, where each node is one generation and edges represent parent-child relationships. Memory is \textbf{shared across all nodes in a lineage but isolated across lineages}. Version structure, node-level outputs, memory state, and inspection records are stored together at the lineage level.

\subsection{Memory Update Rule}

Memory items are represented as tuples over intent, representation, value estimate, and visit count. After each iteration, the stored value is updated by exponential averaging, $q' = (1-\alpha) q + \alpha r$ with $\alpha=0.3$ and reward $r \in [-1,1]$. Retrieval combines semantic similarity and learned value, balancing reuse of relevant past experience against exploitation of historically successful patterns.

\subsection{Why Lineage-Shared Memory}

Two alternatives were considered: per-node memory (each node starts from zero, defeating the purpose of MemRL \cite{zhang2026memrlselfevolvingagentsruntime}) or per-lineage shared memory (all nodes accumulate experience). We chose the latter because (i)~the entire purpose of MemRL is accumulation, (ii)~cross-lineage isolation is preserved at the directory level, and (iii)~a global creativity-rules layer handles truly universal patterns.

\subsection{Three-Layer Architecture}

\emph{Layer~1}: per-lineage learned memory. \emph{Layer~2}: cross-lineage memory resources, including creativity rules, a game pool, and the mechanic archive. \emph{Layer~3}: transient pipeline context for the current generation.

Layer~2 is not merely a passive novelty baseline. The planner queries the global archive for relevant, underexplored, overused, and disfavored mechanics. After reflection, successful generated mechanics can be written back into the archive, creating an initial feedback loop between retrieval and write-back. At the notation level, the archive is best interpreted as an evolving memory over candidate mechanic objects $m$ and partial approximations to their effects on $G_{\text{core}}$.

\section{Runtime Validator}
\label{sec:tester}

\subsection{Motivation}

LLMs frequently produce code that \emph{looks correct} (passes structural keyword checks) but \emph{does not run}. Common failure modes include game-loop functions that are defined but never called, \texttt{requestAnimationFrame} references with no recursive self-call, unbalanced braces, missing canvas context, and DOM access before \texttt{window.onload}.

\subsection{Tier 1: Deep Static Analyzer}

Always runs, no dependencies, $<$10\,ms per game. Performs 9 checks: brace/paren/bracket balance, game loop invocation (\texttt{requestAnimationFrame()} actually called, not just defined), recursive loop self-call, canvas context obtained, input listener attached, init-on-load presence, render-call presence, state-update presence. Each error reduces score by 0.20, each warning by 0.05.

\subsection{Tier 2: Browser Execution Check}

Optional. If browser automation is available, the system launches a headless browser, loads the HTML, waits for canvas paint, sends basic inputs, and collects console errors. Returns $\textit{playable}=\text{True}$ if no console errors occur and the canvas paints successfully. Otherwise, the system degrades gracefully to Tier 1.

\subsection{Pipeline Integration}

The runtime validator is invoked after code generation and before evaluation. If the test fails, a repair stage is invoked with the runtime errors as context, after which the game is re-tested. The runtime score becomes the 7th proxy signal in the reward formula and a hard gate.

\section{Implementation}
\label{sec:implementation}

The current system is implemented as a self-contained Python pipeline. The implementation comprises 6{,}181 lines of Python, excluding generated data, presentation assets, and virtual-environment files. The system directly implements orchestration, memory access, reward computation, runtime validation, mechanic retrieval, lineage recording, and inspection interfaces within a single codebase.

This implementation choice matters because it keeps the full control flow visible and inspectable. The system can expose intermediate mechanic plans, integrate Python-native validation and reward logic, and store lineage records in a format aligned with the analysis questions of this report. The implementation is therefore not just a delivery mechanism for prompts, but part of the experimental contribution itself.

\section{Case Study: Four Real 4-Version Evolutions}
\label{sec:case}

\begin{figure}[htbp]
  \centering
  \includegraphics[width=\linewidth]{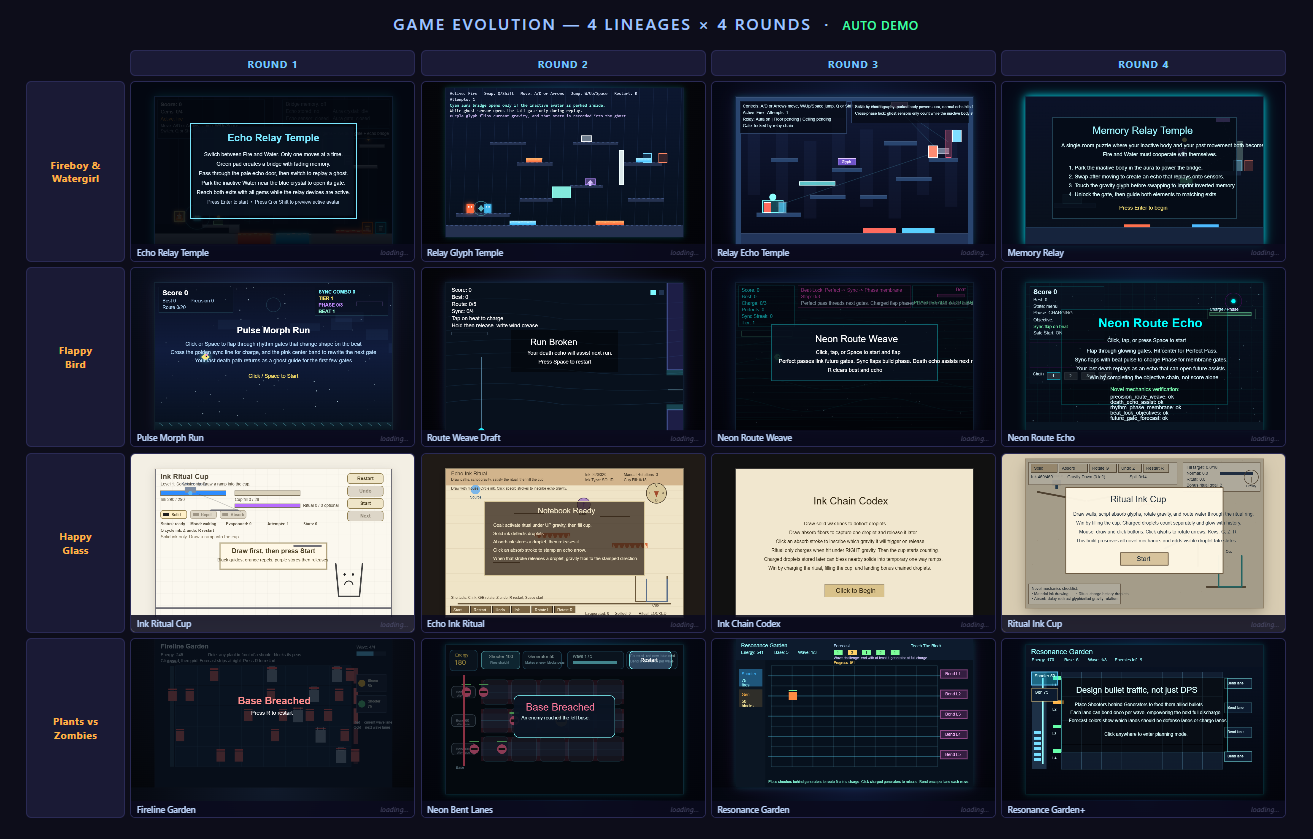}
  \caption{All four 4-round evolution lineages displayed as an auto-demo grid. Each column is one generation round (Round~1--4); each row is one source-game lineage. Games run live in browser with injected demo bots; this screenshot captures representative mid-play states.}
  \label{fig:evolution_anim}
\end{figure}

To illustrate the system in operation, we analyze four real 4-version game demos extracted from the current project website: \textit{Fireboy and Watergirl}, \textit{Flappy Bird}, \textit{Happy Glass}, and \textit{Plants vs.\ Zombies}. Each demo exposes a complete v1--v4 sequence and is useful for a different reason: platform coordination, one-button arcade control, physics-puzzle routing, and lane-defense planning. Together they provide a clearer picture of how the system changes its understanding of a source game across generations. Figure~\ref{fig:evolution_anim} shows all sixteen versions rendered simultaneously in a live browser grid, with each game running an injected demo bot; the screenshot captures representative mid-play states across all four lineages.

\begin{table}[htbp]
\caption{Four representative 4-version evolution sequences. Later versions
typically \emph{reinterpret} the source game around a more explicit mechanic
contract rather than only adding polish.}
\label{tab:case}
\centering
\small
\setlength{\tabcolsep}{8pt}
\renewcommand{\arraystretch}{1.45}
\begin{tabular}{%
  >{\bfseries\raggedright\arraybackslash}p{2.3cm}
  >{\raggedright\arraybackslash}p{4.0cm}
  >{\raggedright\arraybackslash}p{6.5cm}}
\rowcolor{cgNavy}
\color{white}\normalfont Source Game &
\color{white}Initial understanding (v1) &
\color{white}Mechanic reinterpretation (v2–v4) \\[0.5pt]
\rowcolor{cgRowA}
Fireboy \& Watergirl &
Character-switching elemental platform puzzle with relay and timing &
Reinterpreted as \textcolor{cgTeal}{\textbf{memory relay}}: parked bodies,
replay ghosts, and gravity-imprinted echoes become the core puzzle logic \\
\rowcolor{cgRowB}
Flappy Bird &
One-button obstacle dodging with precision timing &
Reinterpreted as \textcolor{cgTeal}{\textbf{route authoring}}: perfect passes
rewrite future gates, death echoes assist later runs, and beat-synced phase
windows change collision logic \\
\rowcolor{cgRowA}
Happy Glass &
Draw-to-route fluid puzzle with physical barriers &
Reinterpreted as \textcolor{cgTeal}{\textbf{programmable fluid logic}}: absorb
strokes store droplets, release events rewrite gravity, and ritual state changes
how fill is counted \\
\rowcolor{cgRowB}
Plants vs.\ Zombies &
Resource-aware lane defense with plant placement and wave management &
Reinterpreted as \textcolor{cgTeal}{\textbf{interception planning}}: generators
intentionally block allied shots, store charge, and discharge through lane
bending / refraction windows \\
\bottomrule
\end{tabular}
\end{table}

\subsection{Fireboy and Watergirl: From Dual-Avatar Traversal to Memory Relay}

The v1 game (\textit{Echo Relay Temple}) already departs from standard dual-character co-op by turning the inactive body into part of the puzzle: one avatar can be parked to power an aura crystal while the active avatar continues traversal. At this stage, however, the game still largely reads as an extended elemental platform puzzle with one additional relay device.

The main shift happens in v2 and becomes clearer in v3--v4. In \textit{Relay Glyph Temple} and \textit{Relay Echo Temple}, swapping no longer functions only as control transfer; it creates a replay ghost that can trigger sensors and open routes. A gravity glyph is then introduced so that the recorded replay inherits a transformed physical rule rather than merely replaying motion. By v4 (\textit{Memory Relay}), the system has a much sharper understanding of the game concept: the puzzle is no longer ``control two elemental bodies'' but ``construct a living circuit out of parked states, ghost replays, and gravity-imprinted memory.'' This is a good example of source-game understanding becoming more mechanic-explicit over time.

\subsection{Flappy Bird: From Obstacle Avoidance to Route Writing}

The v1 flier (\textit{Pulse Morph Run}) still preserves the recognizably simple Flappy Bird interaction loop: one button, vertical impulse, and continuous obstacle traversal. Its main novelty is that gates mutate with rhythm-like timing, so the game is already more structured than a plain obstacle dodger.

The later versions reinterpret what ``passing a gate'' means. In v2 (\textit{Crease Choir}), clean passes can author links into future gates, death traces become ghost echoes that assist later runs, and rhythm timing can charge a temporary phase state. In v3 (\textit{Neon Route Weave}) and v4 (\textit{Neon Route Echo}), this understanding becomes much more coherent: perfect passes are not only scored events, but causal edits to the next route; the last failed run leaves an echo that can open assisted passages; and rhythm is tied to membrane-phasing rather than only cosmetic pulse. The genre understanding therefore shifts from reaction-based survival to a lightweight planning-and-rewrite loop in which the player's past trajectory actively shapes the near future.

\subsection{Happy Glass: From Drawing Supports to Programming Fluid State}

The v1 sequence (\textit{Ink Ritual Cup}) begins close to a recognizable Happy-Glass-style template: the player draws lines to route droplets into a cup while avoiding loss. Even here, the system experiments with multiple ink materials and introduces a ritual checkpoint, but the dominant reading is still ``physics puzzle with drawn barriers.''

By v2 (\textit{Echo Ink Ritual}), absorb strokes can store a droplet and later release it while changing gravity to an inscribed direction. This is the crucial conceptual step: the drawn line becomes not only geometry but a delayed rule trigger. In v3 (\textit{Ink Chain Codex}), the game further adds relay interactions in which charged droplets can bless nearby solids and propagate behavior through neighboring strokes. In v4 (\textit{Ritual Ink Cup}), the overall interpretation becomes cleaner and more legible: solid ink shapes paths, absorb ink scripts delayed state transitions, gravity rotations are limited strategic resources, and ritual-charged droplets are counted separately from normal fill. The source-game idea is therefore re-understood as a small programmable physics language rather than as a pure drawing puzzle.

\subsection{Plants vs.\ Zombies: From Static Lane Defense to Charged Interception}

The v1 lane-defense game (\textit{Fireline Garden}) still reads closest to the original source structure. There are rows, wave previews, plant placement, and lane-local combat. The main novelty is already visible, though: energy generators can physically block allied fire, so resource production and shooting geometry interfere with each other rather than remaining cleanly separated.

That interaction becomes the center of the design in later versions. In v2 (\textit{Neon Bent Lanes}), an entire lane can be bent once per wave, rewriting projectile travel and enemy-path geometry. In v3 (\textit{Resonance Garden}), blocked allied shots are explicitly stored as overcharge in generators and later released as stronger resonance attacks, creating a deliberate ``friendly obstruction as preparation'' mechanic. In v4, this logic is made more strategically readable by forecast-guided lane planning and by framing bends as a once-per-wave refractive discharge window. The understanding of Plants-vs.-Zombies-like play thus shifts from ``place units to stop waves'' to ``plan which lanes should defend directly and which lanes should absorb fire now in order to release stronger refracted attacks later.''

\subsection{Cross-Game Observations}

Figure~\ref{fig:evolution_anim} provides an animated summary of all four lineages side-by-side; the discussion below unpacks each one. Across all four sequences, three common patterns emerge.
\begin{enumerate}[leftmargin=1.5em]
\item The most interesting changes are \textbf{mechanic reinterpretations}, not only visual polish. Later versions tend to re-assign meaning to an existing action: swap becomes memory writing, passing a gate becomes route editing, drawing becomes rule scripting, and blocking allied fire becomes intentional charge storage.
\item The system often moves from \textbf{surface genre mimicry to explicit causal structure}. Early versions preserve the recognizable shell of the source game, while later versions more clearly expose what hidden rule the new variant is really about.
\item The four examples also show that \textbf{game understanding changes by domain}. In platforming, the shift is toward state coordination; in arcade flying, toward future-route shaping; in physics puzzles, toward programmable matter; and in lane defense, toward forecast-based energy planning.
\end{enumerate}

\subsection{Implication for Reward Design}

These case studies illustrate why mechanic-aware records are necessary. If evaluation only asked whether the output still ``looks like'' Flappy Bird or Plants vs.\ Zombies, many of the most interesting changes above would be collapsed into style variation. By contrast, when the system records intended mechanic sets, realized mechanics, and mechanic deltas, it becomes possible to describe evolution in terms of changing rule-bearing structure. This is precisely the level at which the four lineages above become interpretable.

\section{Empirical Results}
\label{sec:results}

\paragraph{Generated data.} The system contains 71 stored lineages: 9 multi-node lineages (up to depth 4) and 62 single-node lineages, for 88 saved nodes overall. The global mechanic archive contains 774 entries, and the summed token count recorded in saved nodes exceeds $4.5 \times 10^6$.

\paragraph{Per-stage computational budget.} The visual generation stage is the largest consumer ($\sim$34\% of total), followed by evaluation ($\sim$27\%), feature generation ($\sim$18\%), skeleton generation ($\sim$9\%), planning ($\sim$8\%), and reflection ($\sim$4\%). The visual stage dominates because it adds presentation detail and animation on top of an already-substantial game body.

\paragraph{Reliability.} After implementing the retry-and-fallback mechanisms in Section~\ref{sec:system}, pipeline success rate is $>$98\%, with empty-output recovery rate $>$95\% within 3 retries.

\paragraph{Score distribution.} Across all generated games' final iteration, average creativity is $\sim$7.0/10, average evaluator \textit{playability} score is $\sim$6.5/10, and average overall is $\sim$6.2/10. The terminology here follows the current evaluator schema, but the field should be interpreted as a coarse proxy for usability and functional completeness rather than as a validated measure of player enjoyment. These scores are also subject to LLM scoring saturation (Section~\ref{sec:reward}) and are not validated against human judgment.

\section{Evaluation Protocol}
\label{sec:protocol}

\paragraph{Prompt dataset.} Game prompts were drawn from an internal prompt library spanning multiple genre categories. Each prompt was used to generate one lineage, with up to three internal refinement iterations.

\paragraph{Representative source games for mechanic coverage.} For simple, legible mechanic exemplars, the current project also maintains a strict reference table of 252 web games with compressed genre and tag vocabularies. Four especially useful anchors for report discussion are \textit{Flappy Bird} (single-input reaction loop), \textit{Fireboy and Watergirl} (dual-character co-op platforming), \textit{Happy Glass} (draw-to-shape physics puzzle), and \textit{Plants vs.\ Zombies} (resource-aware lane defense). These examples are useful not because the system clones them directly, but because they span very different mechanic structures while remaining simple enough to explain and retrieve.

\paragraph{Generation settings.} Maximum iterations 3; planning temperature 0.7; evaluation temperature 0.2; runtime validation enabled; optional browser-based execution check; memory retrieval top-k = 5.

\paragraph{Measurement definitions.}
\begin{itemize}[leftmargin=1.5em,topsep=0.2em]
  \item \emph{Generation time} is measured from the start of planning to the completion of final formatting, including any internal repair loops.
  \item \emph{Computational usage} includes all model consumption from planning, generation, evaluation, and reflection, including retries and fallback executions.
  \item \emph{Evaluator playability} follows the current evaluator field name, but operationally it should be read as a coarse usability/completeness proxy. Runtime robustness is measured separately by the validator score combining static analysis and (if available) browser execution.
\end{itemize}

\paragraph{Model configuration.} The system uses contemporary large-language-model backends. Some lineages were generated with Kimi (server) and others with GPT-class models (local). Python 3.12, Ubuntu 24.04 (server) / Windows 11 (development).

\section{Related Work and Discussion}
\label{sec:related}

\paragraph{Multi-agent code generation.} Frameworks such as ChatDev, MetaGPT, and AgentVerse decompose software generation into role-based agents \cite{chatdev,hong2024metagpt,chen2023agentverse}. CreativeGame follows the general intuition of role specialization, but organizes it around a fixed iterative pipeline for game generation, testing, evaluation, reflection, and memory writing.

\paragraph{LLM creativity evaluation.} Prior work on creativity theory and assessment emphasizes both the difficulty of defining creativity beyond simple novelty and the importance of who gets to judge it \cite{KaufmanBaer2012,GlaveanuBeghetto2021}. More recent work on LLM-as-a-judge has shown both the promise and the limitations of model-based evaluation in open-ended settings \cite{zheng2023judging}. Our CreativeProxyReward differs from judge-heavy LLM evaluation by minimizing the LLM judgment component to a single auxiliary signal and grounding the primary signals in deterministic Python computations.

\paragraph{Formal game and mechanic structure.} Relative to game-studies-inspired formalizations, the present project adopts a structural view of games as rule-organized systems $G$ and mechanics as local rule structures $m$. This framing is important because it clarifies what should count as creativity in the pipeline: changes to $G_{\text{core}}$ rather than merely to $G_{\text{support}}$, and mechanic-level changes rather than purely thematic or cosmetic variation. In this respect, the report is closest to game-design accounts that foreground rules, interaction, and meaningful consequence \cite{SalenZimmerman2003,Schell2008}.

\paragraph{Game design foundations.} Foundational game-design texts consistently stress the interaction between rules, player understanding, iteration, and the design of meaningful experience \cite{SalenZimmerman2003,Schell2008,Koster2005}. The present project extends these concerns into an automated setting by making mechanic planning, realization, and cross-version change explicit objects inside the generation loop.

\paragraph{Memory-augmented agents.} Memory-augmented agent systems introduce persistent experience into sequential decision processes. Recent MemRL-style work makes this idea explicit through runtime reinforcement learning over episodic memory \cite{zhang2026memrlselfevolvingagentsruntime}. We adapt this general direction to creative generation, with the design choice of \emph{lineage-scoped} sharing rather than per-task isolation, motivated by the desire for cross-version experience accumulation.

\paragraph{Runtime validation in code generation.} Evaluation work on code-generating language models has strongly emphasized execution-based correctness \cite{chen2021evaluatingcode}. Our innovation is integrating runtime validation as both a reward signal and a repair trigger within the multi-agent pipeline, with a graceful degradation path when richer execution checks are unavailable.

\section{Conclusion}
\label{sec:conclusion}

We presented CreativeGame, a multi-agent system for iterative creative game generation. Its central contributions are a CreativeProxyReward whose primary signals are programmatic, a lineage-aware memory that enables cross-version experience accumulation, runtime validation integrated as both a reward signal and a hard gate, and a mechanic-guided planning layer in which retrieved archive knowledge is converted into an explicit mechanic plan. The real lineage case study demonstrates that version-to-version mechanic evolution can be recorded, inspected, and discussed in concrete structural terms.

Taken together, the system shows that creativity in game generation can be approached as an inspectable engineering problem: mechanics can be planned explicitly, evaluated structurally, stored across generations, and followed through iterative evolution rather than treated only as a final subjective impression.

\bibliographystyle{IEEEtran}
\bibliography{creativegame_refs}

\end{document}